\documentclass[journal]{IEEEtran}
\ifCLASSINFOpdf
\else
\fi
\label{begins}

\usepackage{graphicx}
\usepackage{subfig}
\usepackage{cite}
\usepackage{bm}
\usepackage{algorithm}
\usepackage[noend]{algpseudocode}
\usepackage{multicol,multirow,booktabs}
\usepackage{makecell}
\usepackage{amsmath}
\usepackage{longtable}
\usepackage{rotating}
\usepackage{caption}
\usepackage{amssymb}
\usepackage{pifont}
\usepackage{array}
\usepackage{color}
\usepackage{lineno}
\usepackage{setspace}

\usepackage[capitalize]{cleveref}
\crefname{section}{Sec.}{Secs.}
\Crefname{section}{Section}{Sections}
\Crefname{table}{Table}{Tables}
\crefname{table}{Tab.}{Tabs.}

\begin{document}

%
\title{Full-attention based Neural Architecture Search using Context Auto-regression}
%
%
%

\author{Yuan~Zhou,
        Haiyang~Wang, Shuwei~Huo~and Boyu~Wang
\thanks{Yuan Zhou, Haiyang Wang and Shuwei Huo are with the School of Electrical and Information Engineering, Tianjin University, Tianjin 300072, China.}
\thanks{Boyu Wang is with the Department of Computer Science, University of Western Ontario, Canada.}}
%
%

\markboth{}%
{Shell \MakeLowercase{\textit{\emph{et al.}}}: Bare Demo of IEEEtran.cls for IEEE Journals}

%



\maketitle

\begin{abstract}
   Self-attention architectures have emerged as a recent advancement for improving the performance of vision tasks. Manual determination of the architecture for self-attention networks relies on the experience of experts and cannot automatically adapt to various scenarios. Meanwhile, neural architecture search (NAS) has significantly advanced the automatic design of neural architectures. Thus, it is appropriate to consider using NAS methods to discover a better self-attention architecture automatically. However, it is challenging to directly use existing NAS methods to search attention networks because of the uniform cell-based search space and the lack of long-term content dependencies. To address this issue, we propose a full-attention based NAS method. More specifically, a stage-wise search space is constructed that allows various attention operations to be adopted for different layers of a network. To extract global features, a self-supervised search algorithm is proposed that uses context auto-regression to discover the full-attention architecture. To verify the efficacy of the proposed methods, we conducted extensive experiments on various learning tasks, including image classification, fine-grained image recognition, and zero-shot image retrieval. The empirical results show strong evidence that our method is capable of discovering high-performance, full-attention architectures while guaranteeing the required search efficiency. 

\end{abstract}


%
\IEEEpeerreviewmaketitle

\section{Introduction}

The design and construction of neural network architectures are critical concerns because better network architectures usually lead to significant performance improvements. However, deep neural architectures often require elaborate design for specific tasks, which indicates that designing a suitable architecture requires tremendous effort from human experts. To eliminate such extensive engineering, neural architecture search (NAS) methods \cite{1,2,5,7} have been proposed to automate the design of neural architectures. Many architectures produced by NAS methods have achieved higher accuracy than those manually designed for tasks, such as image classification \cite{1,7}, semantic segmentation \cite{14,15}, and object detection \cite{16}. NAS methods boost the model's performance, and free human experts from the tedious task of tweaking the architecture. A trend has been growing towards automatically designing neural network architectures instead of relying on human effort and experience.

Recently, self-attention network design \cite{27,29,30} has made significant progress. Content-based interactions and the ability to capture long-term dependencies have made self-attention a critical component in neural networks. Ramachandran et al.\cite{27} propose a pure self-attention vision model that replaces every spatial convolution with a self-attention operator. Zhao et al. \cite{30} explore variations in the self-attention operator and assess the effectiveness of image recognition models that are based fully on self-attention. These studies show that self-attention operations can be the basic building block to build image recognition models. Although the development of self-attention network designs has made significant progress, manually designing appropriate full-attention architectures is still a challenging task that requires substantial efforts by human experts, especially as the number of design choices increases. There is increasing interest in the automatic design of neural network architectures, rather than relying on the knowledge and experience of human experts. It is essential to consider using NAS methods to pursue better a self-attention architecture design. 

Most existing NAS methods are designed to search convolutional neural networks, but it is challenging to adopt these methods to search for full-attention networks. First, these methods \cite{2,4,7} usually adopt cell-based search spaces, where the cell structures in the shallow and deep layers of the network are identical. This is not suitable for self-attention structures because self-attention plays different roles in different stages of the network. Second, existing NAS methods \cite{1,11,13} generally use classification tasks as the supervision of architecture search and guide the optimization of network structures using classification accuracy as the evaluation criterion. The classification task requires the model to pay more attention to extracting features from the local region related to the classification label without considering long-term dependency. Nevertheless, self-attention models focus on capturing long-term content dependencies between pixels to learn rich and broadly transferable representations. Therefore, search methods that use classification tasks as the supervision are not suitable for searching attention structures. 

To address these issues, this paper proposes a novel NAS method to design full-attention architectures automatically. First, a stage-wise search space is established that allows the search algorithm to choose different self-attention operations for each stage of the network more flexibly. Moreover, a new self-supervised task is designed based on context auto-regression which reconstructs local missing content by integrating it with the global feature. Using the proposed context auto-regression as supervision, long-term dependencies across pixels can be extracted without requiring the images' label information. 

Specifically, based on the designed self-supervised task, we propose a task-wise search method. The method includes a search phase and a fine-tuning phase with different tasks being used to supervise the architecture search in different phases. In the search phase, the designed self-supervised task is used as the supervision, encouraging the self-attention model to capture long-term dependencies. In the fine-tuning stage, classification task is used as the supervision to search for self-attention network structures.

The main contributions of this paper are summarized as follows: 

1)	A full-attention based neural architecture search method is proposed that automatically designs networks that use self-attention operations as the primary building block. 

2)	A stage-wise search space is established in which different self-attention operations can be selected for each stage of the network.

3)	A new self-supervised task based on context auto-regression is designed and used to train a self-supervised search algorithm. Compared to supervised learning-based search, the proposed self-supervised search enables the capture of long-range dependencies, thus constructing better full-attention networks. 

4)	Extensive experiments demonstrate that our method can discover high-performance full-attention networks for image classification, fine-grained image recognition and zero-shot image retrieval.


\section{Related Work}
\label{sec:intro}

\textbf{Neural Architecture Search.} Recently, NAS which aims to design neural networks automatically has attracted increasing attention. Existing NAS approaches can be roughly divided into three categories, namely, reinforcement learning (RL)-based methods \cite{1,2,3,4}, evolutionary algorithms (EA)-based methods \cite{5,6,10}, and gradient-based methods \cite{7,11,12,13}. RL-based methods train a recurrent neural network as a controller to generate a series of actions to specify the CNN architecture. An alternative search technique is to use evolutionary algorithms, that ''evolve'' the architecture by mutating the optimal architectures to find a neural architecture tailored for a given task. Although these works achieved state-of-the-art results on various classification tasks, their main disadvantage was that they demand excessive computational resources. 

To alleviate this issue, gradient-based NAS methods have been proposed to speed up the searching process. In contrast to treating architecture search as a black-box optimization problem, gradient-based NAS methods relax the discrete and non-differentiable architecture representations to be continuous, so that the gradient obtained during the training process can be used to optimize the architecture. Therefore, gradient-based methods successfully accelerate the architecture search process, and usually require only a few GPU days. In addition, existing methods for CNN usually require the model to extract features from the local region related to the classification label, without considering long-term dependency. In contrast to these methods, the proposed search method is enabled to capture long-range dependencies, thus constructing better full-attention networks.

\textbf{Self-attention mechanism.} Self-attention mechanisms have been widely adopted in various tasks, such as machine translation \cite{17}, generative modeling \cite{18}, and visual recognition \cite{19,20,21}. Vaswani et al. \cite{17} proposed the Transformer, which is one of the first attempts to apply a self-attention mechanism to model long-range dependencies in machine translation. Wang et al. \cite{19} extended a sequential self-attention network to a spacetime non-local network to capture long-range dependencies in videos. The proposed non-local block significantly improved the video classification accuracy of CNNs. Beyond assisting CNNs to deal with long-range dependencies, Ramachandran et al. \cite{27} developed a local self-attention module to limit the amount of computation and leverage this module to build a fully attentional vision model. In addition, \cite{28,29,30} also proposed different forms of self-attention to construct pure-attention vision models. In recent work, vision-transformer-based methods \cite{31,32,33} have attracted considerable interest. They explored a self-attention architectural design that effectively learns visual representation.

Different from these methods for manually designing self-attention models, we propose a NAS algorithm that automatically discovers the optimal self-attention network. Furthermore, the discovered network is demonstrated to achieve superior performance.

\begin{figure*}[t]
	\centering
	\includegraphics[width=0.9\linewidth]{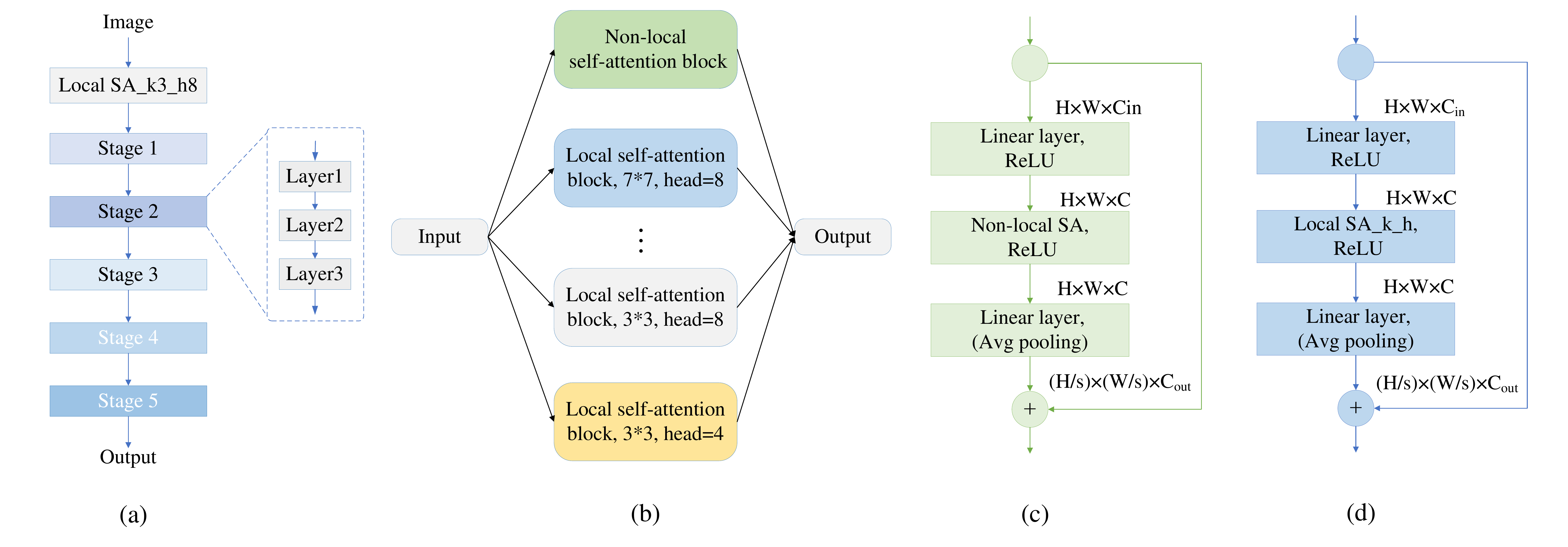} 
	\caption{Illustration of the search space. (a) The macro-architecture consists of five stages, each stage having three intermediate searchable layers. For simplicity, only the structure of the stage 2 is shown. (b) Each layer contains different candidate self-attention blocks. (c) The structure of the non-local self-attention block. (d) The structure of the local multi-head self-attention block.}.
	\label{fig1}
\end{figure*}

\section{Full-attention based Neural Architecture Search}

In this section, a novel NAS method is proposed to address the challenges of designing a full-attention network. First, a flexible stage-wise search space is proposed that allows various attention operations to be used for different stages of the network. Figure 1 illustrates the search space for the proposed method. Then, a continuous relaxation of the discrete architectures is developed, such that the architecture can be optimized by gradient descent. Moreover, a self-supervised search method using context auto-regression is proposed to discover a full-attention architecture. 

\subsection{Search Space}

In this study, we propose a stage-wise search space to search for a full-attention network. \Cref{fig1}(a) shows the macro-architecture of the search space. The searched model is partitioned into a sequence of pre-defined stages that gradually reduce spatial resolutions of the feature maps. The precise architecture configurations are listed in \Cref{table1}. The macro-architecture defines the number of layers, input dimensions, and output channel number for each layer. It also specifies which layers of the network are to be searched. The first and last layers of the network have fixed operators. The remainder of the macro-architecture consists of five stages, and each stage has three intermediate searchable layers.

\begin{table}[]
	\centering
	\renewcommand{\arraystretch}{0.9}
	\renewcommand\tabcolsep{2.0pt} 
	\begin{tabular}{c|c|c|c|c|c}
		\hline
		& Input shape & Operations       & Channels & n & s \\ \hline
		Stem          & 32$\times$32$\times$3     & Local SA\_k3\_h8 & 16       & 1       & 1 \\
		Stage 1       & 32$\times$32$\times$16    & To Be Searched   & 16       & 3       & 1 \\
		Stage 2       & 32$\times$32$\times$16    & To Be Searched   & 32      & 3       & 2 \\
		Stage 3       & 16$\times$16$\times$32   & To Be Searched   & 32      & 3       & 1 \\
		Stage 4       & 16$\times$16$\times$32   & To Be Searched   & 64      & 3       & 2 \\
		Stage 5       & 8$\times$8$\times$64     & To Be Searched   & 64      & 3       & 1 \\
		Pooling layer & 8$\times$8$\times$64     & Avgpool          & 64      & 1       & 1 \\
		Output        & 1$\times$1$\times$64     & FC               & 10       & 1       & 1 \\ \hline
	\end{tabular}
	\caption{Macro-architecture of the search space. ''Operations'' denotes the type of operation block. ''channels? denotes the output channel number. ''n'' denotes the number of intermediate searchable layers. ''s'' denotes the stride of the first layer in a stage.}
	\label{table1}
\end{table}

Two types of self-attention operations i.e., local multi-head self-attention \cite{27} and non-local self-attention \cite{19}, are considered as candidate operations in a stage. Local multi-head self-attention extracts local spatial information and multiple heads are used to learn multiple distinct representations of the input, analogous to group convolutions \cite{61,62}. The size of the local spatial extent  space and the number of self-attention heads are important settings in the local multi-head self-attention. A non-local self-attention operation \cite{19} can be regarded as a global context modeling module that explicitly capturing long-range interactions among distant positions. In summary, the candidate operations consist of seven self-attention operations, and their configurations are listed in \Cref{table2}. We allow the selection of different operations, including non-local self-attention and local self-attention, with different spatial extents and different head numbers.

Based on these candidate self-attention operations, two types of building blocks are designed as candidate blocks of the macro-architecture. Each searchable layer in the macro-architecture can choose a block. The non-local building block is illustrated in \Cref{fig1}(c) and the local building block is illustrated in \Cref{fig1}(d). The block structure is inspired by the ''bottleneck'' building block design \cite{39}. 

As shown in \Cref{fig1}(c), the non-local building block contains a linear layer, followed by a non-local self-attention operation, followed by another linear layer. The first linear layer transforms the input features and reduces their channel dimensionality for efficient processing. The final linear layer expands the features to match the output's dimensionality. If the stride of the layer is set to two, there is an average pooling operation after the final linear layer to reduce the spatial resolution. The ''ReLU'' activation functions follow the first linear layer and the attention operation. In addition, the shortcut connections simply perform identity mapping, and their outputs are added to the outputs of the stacked layers. 

As shown in \Cref{fig1}(d), the local building block has a similar structure, except that the intermediate non-local self-attention operation is replaced by a local self-attention operation with different spatial dimensions and head numbers.

In summary, our overall search space contains 15 searchable layers and each layer can choose from seven candidate blocks, thus producing  $7^{15}$  possible architectures. Finding the optimal network structure from such an enormous search space is a non-trivial task. 

\begin{table}[]
	\centering
	\renewcommand{\arraystretch}{0.9}
	\renewcommand\tabcolsep{9.0pt}

	\begin{tabular}{c|c|c}
		\hline
		Operations       & Spatial Extent & Head \\ \hline
		Local SA\_k3\_h4 & 3$\times$3            & 4    \\
		Local SA\_k3\_h8 & 3$\times$3            & 8    \\
		Local SA\_k5\_h4 & 5$\times$5            & 4    \\
		Local SA\_k5\_h8 & 5$\times$5            & 8    \\
		Local SA\_k7\_h4 & 7$\times$7            & 4    \\
		Local SA\_k7\_h8 & 7$\times$7            & 8    \\
		Non-local SA     & -              & -    \\ \hline
	\end{tabular}
	\caption{Configurations of candidate operations in the search space}
	\label{table2}
\end{table}

\subsection{Differentiable Formulation of the Search Space}

For the search phase, it is infeasible to solve the search problem through brute-force enumeration of the search space. In this work, we use the differentiable architecture search method in \cite{7} to efficiently find the optimal self-attention network. We develop a differentiable formulation for our proposed search space. The formulation enables the architecture and its weights to be jointly optimized with back-propagation, thus reduces the search time significantly compared to explicitly sampling and evaluating different architectures.

\begin{figure}[t]
	\centering
	\includegraphics[width=0.9\columnwidth]{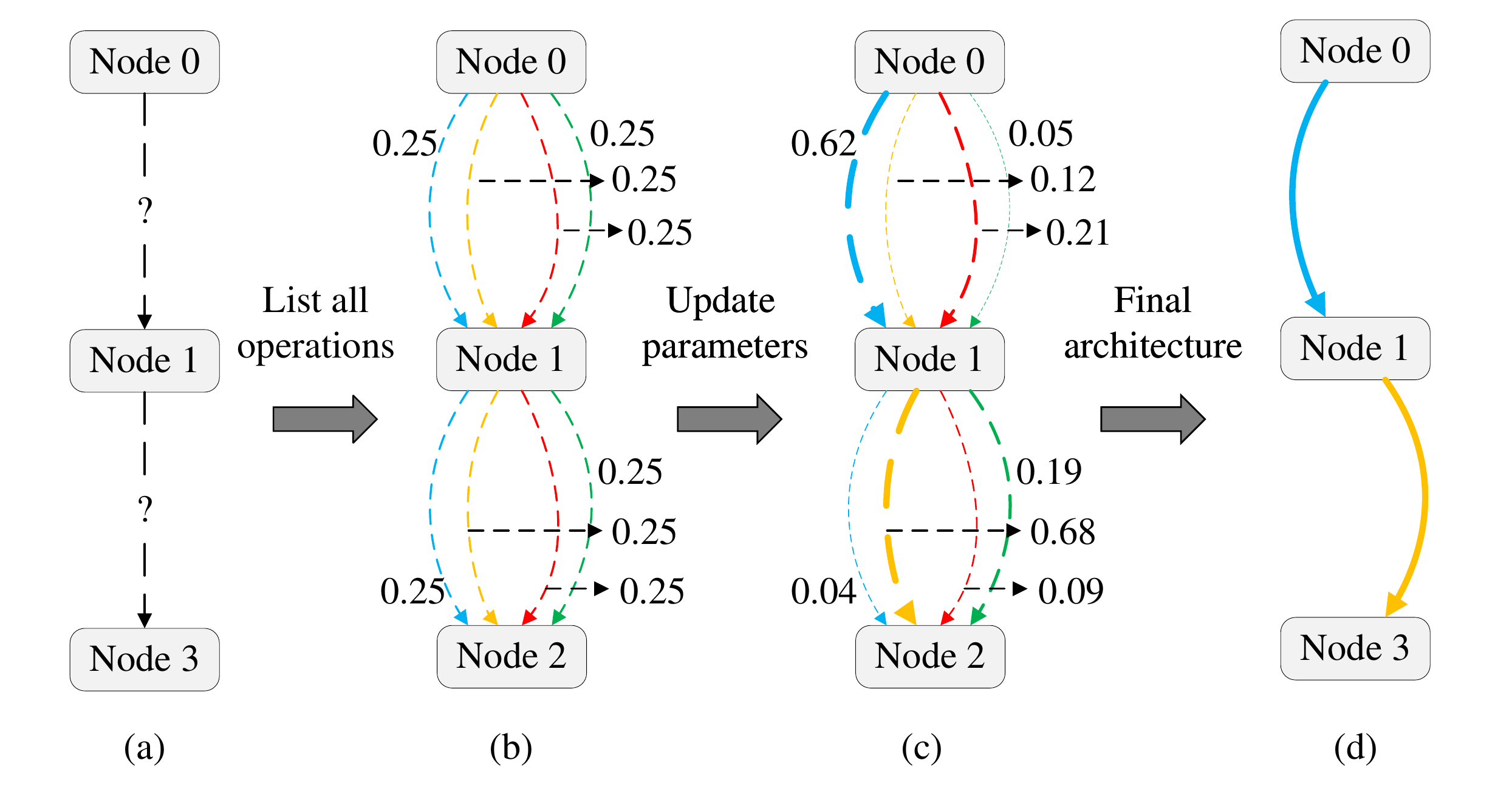} 
	\caption{Illustration of the differentiable architecture search procedure. (a) The search space is represented as a directed acyclic graph. The node represents the feature map and the edge represents an unknown operation. (b) All operations between each edge (shown by connections with different colors) are listed. The architecture parameters are shown next to each connection. (c) During the search phase, the architecture parameters are constantly updated. (d) At the end of the search phase, a stand-alone architecture is obtained based on the architecture parameters.}.
	\label{fig3}
\end{figure}

The overall procedure for the differentiable architecture search is shown in \Cref{fig3}. The search space is represented as a directed acyclic graph of N nodes, where each node represents a feature map. We denote the operation space as $\mathcal{O}$, where each element represents a candidate operation $b()$. The edges $f_{i,j}$ represents the information flow connecting node $i$ and node $j$, which consists of a set of operations weighted by the architecture parameters $\alpha _{(i,j)}$, and is thus formulated as:
\begin{equation}
	{f_{i,j}}\left( {{x_i}} \right) = \sum\limits_{b \in \mathcal{O}} {\frac{{\exp \left( {\alpha _{i,j}^b} \right)}}{{\sum\nolimits_{b' \in \mathcal{O}} {\exp \left( {\alpha _{i,j}^{b'}} \right)} }}} b\left( {{x_i}} \right)
\end{equation}
where the operation weights for a pair of nodes are parameterized by a vector $\alpha _{(i,j)}$ of dimension $\left| \emph{O} \right|$.

After relaxation, the architecture parameters $\alpha$ and the weight parameters $w$ (e.g., the weights of the self-attention operations) can be optimized by applying alternating gradient descent performed on the training and validation sets. This optimization procedure is defined as a bilevel optimization problem:

\begin{equation}
	\mathop {\min }\limits_\alpha  {L_{valid}}\left( {\alpha ,{w^ * }\left( \alpha  \right)} \right)
\end{equation}
\begin{equation}
	s.t. \  {\rm{ }}{w^ * }\left( \alpha  \right) = \mathop {\arg \min }\limits_w {L_{train}}\left( {\alpha ,w} \right)
\end{equation}

where ${L_{valid}}$ is the validation loss and ${L_{train}}$ is the training loss. Once the training is completed, a discrete architecture can be obtained by replacing each mixed operation block with the most probable block, that is, ${f_{i,j}} = \mathop {\arg \max }\limits_{b \in O} \alpha _{i,j}^b$.

\begin{figure}
	\centering
	\includegraphics[width=0.9\columnwidth]{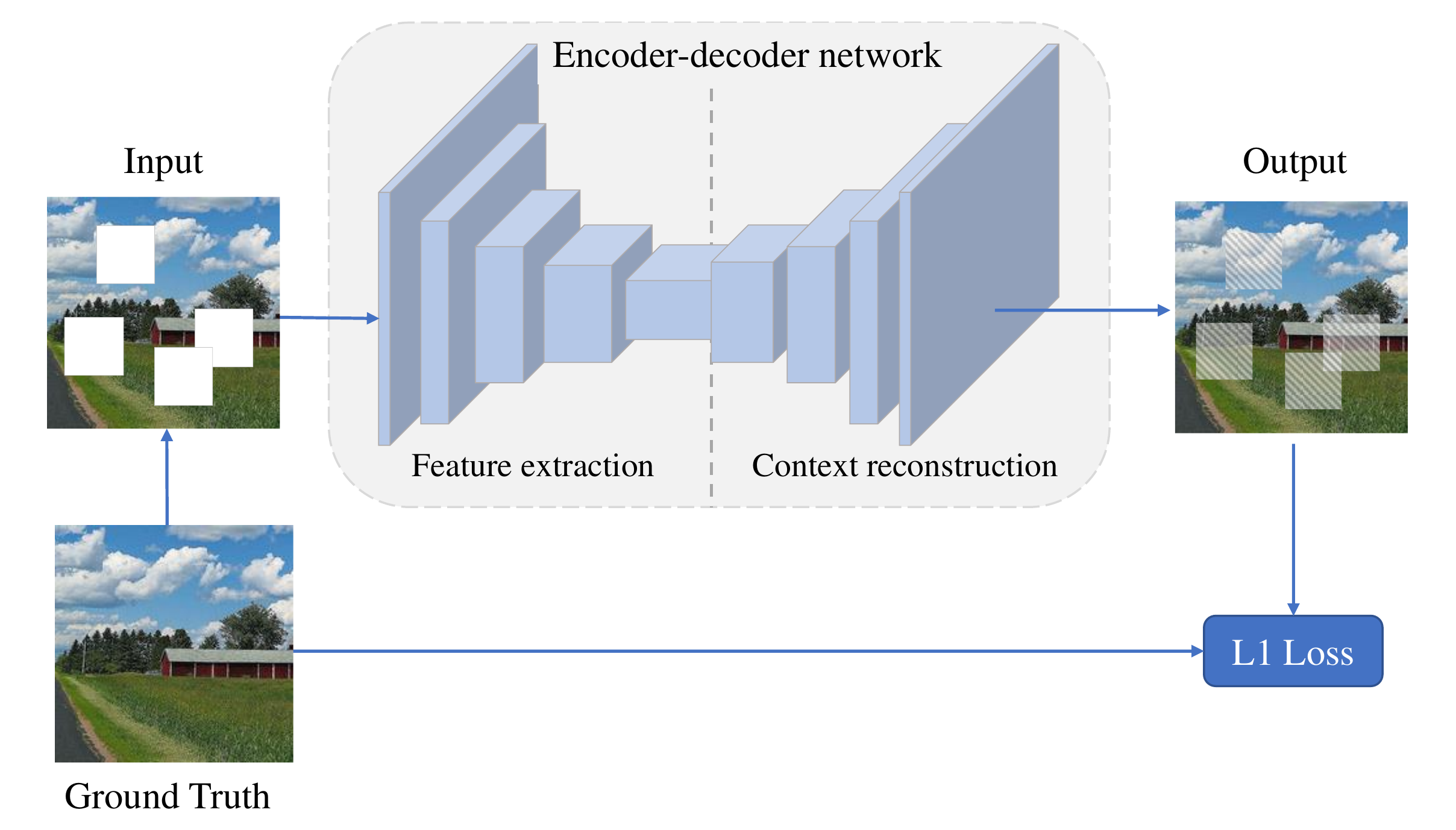} 
	\caption{ Illustration of the context auto-regression. Several regions on the input image are randomly masked before it was fed into an encoder-decoder network. The network extracts feature and reconstructs the missing image content. }.
	\label{fig4}
\end{figure}

\subsection{Context Auto-regression based search method} 
Current NAS methods typically use classification tasks for the supervision of architecture search. However, the classification task requires the model to fix more attention on extracting features from the local area related to the classification label, without considering the long-term dependence. In this case, the current search methods are not suitable for searching self-attention structures that focus on capturing the long-term content dependencies between pixels. A new self-supervised task, termed context auto-regression, is designed in this study. Based on this task, a self-supervised search method is proposed to discover optimal self-attention architecture. The proposed search method using context auto-regression encourages the modeling of more global dependencies, which is more suitable for searching full-attention networks.

Context auto-regression is now proposed. As shown in \Cref{fig4}, given an unlabeled image, several regions are randomly marked on the input image before it is fed into an encoder-decoder network. The encoder then takes the input image with missing regions and produces a feature representation of the image. The decoder uses this feature representation and reconstructs the missing image content. The network is trained by regressing the ground truth content. L1 loss is used as the reconstruction loss to capture the overall structure of the missing region and coherence regarding its context. Instead of choosing a single mask at a fixed location, several smaller, possibly overlapping, random masks are added that cover up to 1/4 of the image. To succeed in this task, the model has to understand the content of the entire image, as well as produce a plausible hypothesis for the missing parts.

Based on this task, a self-supervised search method is proposed for learning the full-attention architecture. The macro-architecture shown in \Cref{table2} is used as the encoder to extract the features. Then, the feature is passed through a series of up-sampling layers to produce the prediction. During the search phase, we jointly learn the architecture parameters and weight parameters of the macro-architecture using context auto-regression.

\begin{figure*}[t]
	\centering
	\includegraphics[width=0.9\linewidth]{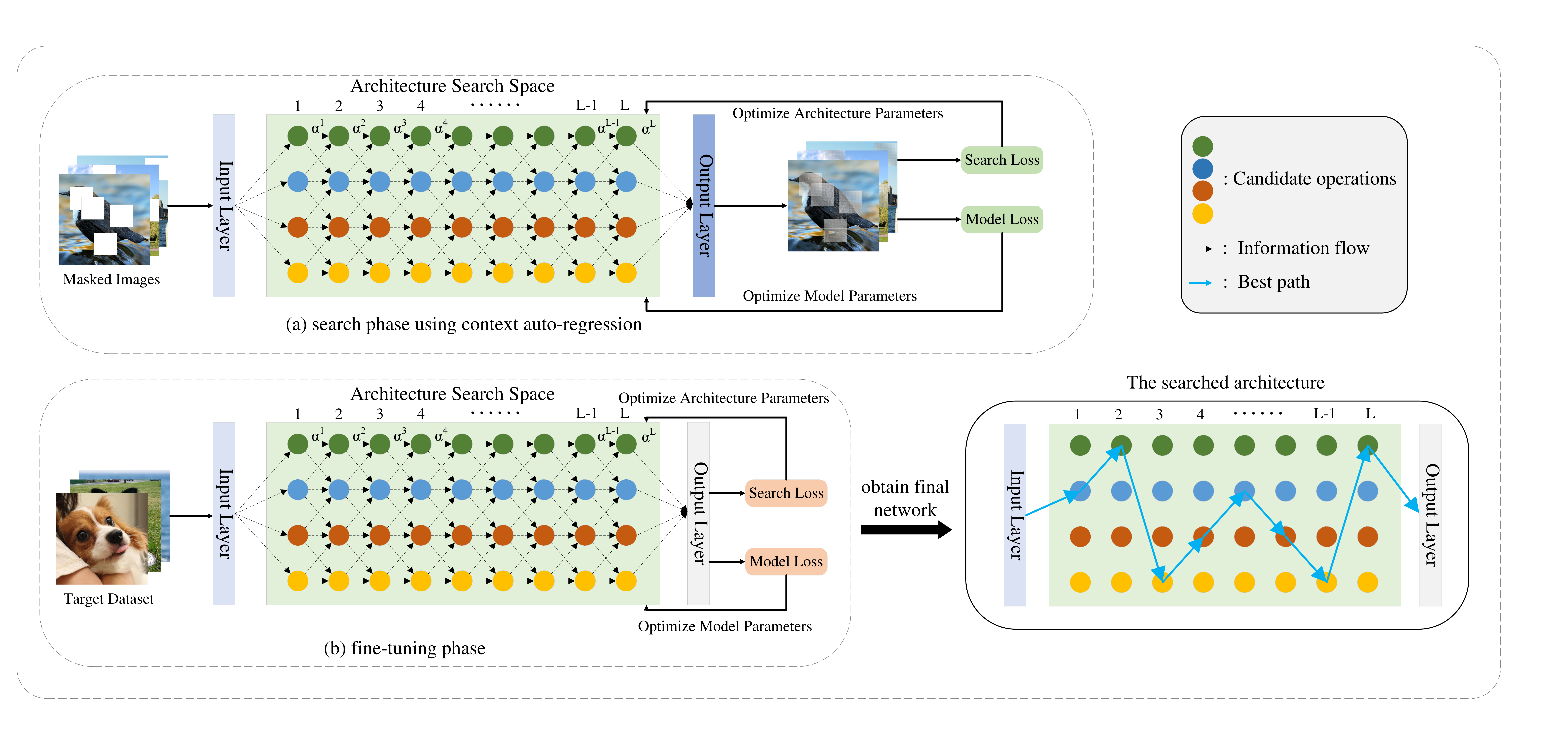} 
	\caption{Illustration of our search procedure. The procedure consists of a search phase followed by a fine-tuning phase. In the search phase, we propose a self-supervised algorithm applying context auto-regression to guide the architecture search. When fine-tuning, we perform the search on target tasks to obtain the final architecture.}.
	\label{fig5}
\end{figure*}

\subsection{Fine-tuning}
Our method consists of a search phase using context auto-regression followed by a fine-tuning phase. \Cref{fig5} illustrates the overall procedure of the proposed method. In the search phase, context auto-regression is used to learn the architecture parameters of the search network. When the search phase is complete, we store the architecture parameters and use them as initialized values for the fine-tuning phase. When fine-tuning, a differentiable architecture search is performed on the target tasks. Here we take the classification task as an example. Following \cite{7}, the training data are partitioned into two separate sets for training and validation. First-order approximation is adopted while the architecture parameters $\alpha$ and weight parameters $w$ are alternately optimized using gradient descent. Specifically, in an iterative manner, the weight parameters are optimized by descending $\bigtriangledown_{w}L_{train}(\alpha, w)$  on the training set, and the architecture parameters by descending $\bigtriangledown_{\alpha}L_{val}(\alpha, w)$ on the validation set. When the fine-tuning procedure is completed, we obtain the final architecture according to the architecture parameters.

\section{Experiments}
In this section, we perform experiments on image classification, fine-grained image recognition, and zero-shot image retrieval tasks, and compare the performance of each proposed models with other state-of-the-art models.

\subsection{Datasets}

The experiments are first conducted on three popular image classification datasets, namely CIFAR10 \cite{34}, CIFAR100, and ImageNet \cite{35}. Then a fine-grained visual categorization dataset, the Caltech-UCSD Birds dataset (CUB-200-2011) \cite{36} is used to test the transferability of the architectures discovered on CIFAR10.

CIFAR10 and CIFAR100 contain 50 K training and 10 K testing RGB images with a fixed spatial resolution of 32$\times$32. CIFAR-10 categorizes images into 10 classes, whereas CIFAR-100 has 100 classes. The ImageNet 2012 dataset consists of 1.28 million training images and 50 K validation images from 1000 different classes. CUB-200-2011 has 11,788 images representing 200 bird categories that are split into 5,994 training and 5,794 test images.
\begin{table*}[]
	\centering
	\renewcommand{\arraystretch}{0.8}
	\renewcommand\tabcolsep{4pt} 
	\begin{tabular}{ccccccc}
		\hline
		\multirow{2}{*}{Model}      & \multirow{2}{*}{Type}            & \multirow{2}{*}{Design method} & \multirow{2}{*}{\begin{tabular}[c]{@{}c@{}}Search cost\\ (GPU days)\end{tabular}} & \multirow{2}{*}{Params} & \multirow{2}{*}{\begin{tabular}[c]{@{}c@{}}Error on\\ CIFAR-10\end{tabular}} & \multirow{2}{*}{\begin{tabular}[c]{@{}c@{}}Error on\\ CIFAR-100\end{tabular}} \\
		&                                  &                                &                                         &                         &                                    &                                     \\ \hline
		SENet \cite{24}              &    conv-based                               & manual                         & -                                       & 11.2M                   & 4.05\%                             & -                                   \\
		Wide ResNet \cite{41}           &     conv-based                              & manual                         & -                                       & 8.9M                   & 4.53\%                             & 21.18\%                             \\
		ResNeXt-29 \cite{62}           &    conv-based                               & manual                         &-                                         & 34.4M                   & 3.65\%                             & -                                   \\
		UPANets32 \cite{42}          &     conv-based                              & manual                         & -                                       & 5.93M                   & 4.12\%                             & 21.22\%                             \\
		SimpleNet V2 \cite{43}       &      conv-based                             & manual                         & -                                       & 8.9M                    & 4.11\%                             & 20.83\%                             \\ \hline
		MetaQNN \cite{44}            & conv-based       & auto                             & 80                                      & 11.2M                   & 6.92\%                             & 27.14\%                                   \\
		NAS \cite{1}              &            conv-based                       & auto                             & 16000                                   & 7.1M                    & 4.47\%                             & -                                   \\
		Net Transformation \cite{45} &       conv-based                            & auto                             & 10                                      & 19.7M                   & 5.70\%                             & -                                   \\
		SMASH \cite{46}             &         conv-based                          & auto                             & 1.5                                     & 16M                     & 4.03\%                             & 22.07\%                                   \\
		Hierachical NAS \cite{6}     &        conv-based                           & auto                      & 300                                     & 61.3M                   & 3.63\%                             & -                                   \\
		NSGANet \cite{47}            &          conv-based                         & auto                      & 8                                       & 3.3M                    & 3.85\%                             & 20.74\%                                   \\
		ENAS \cite{4}    &          conv-based                         & auto                           & 0.32                                     & 38M                    & 3.87\%                             & -                                   \\ \hline
		NesT-T \cite{33}             & attention-based & manual                         & -                                       & 6.2M                    & 3.96\%                             & 21.31\%                             \\
		DeiT-B \cite{49}            &      attention-based                            & manual                         & -                                       & 85.1M                   & 7.59\%                             & 29.51\%                             \\
		PVT-S \cite{50}              &       attention-based                           & manual                         & -                                       & 24.1M                   & 7.66\%                             & 30.21\%                             \\
		CCT-6/3$\times$1 \cite{51}          &       attention-based                           & manual                         & -                                       & 3.17M                   & 4.71\%                             & 22.69\%                             \\
		Swin-B\cite{52}             &        attention-based                          & manual                         & -                                       & 86.7M                   & 5.45\%                             & 21.55\%                             \\ \hline
		Our method           &     attention-based                             & auto                 & 2                                       & 2.52M                   & 3.44\%                             & 20.68\%                             \\ \hline
	\end{tabular}
	\caption{Classification errors on CIFAR-10 and CIFAR-100}
	\label{table3}
	
\end{table*}

\subsection{Implementation Details}

The experiments consist of three phases. First, a subset of ImageNet is searched using context auto-regression. Then, based on the architecture parameters obtained in the first phase, fine-tuning is performed on CIFAR-10 to obtain the optimal self-attention network. Finally, the searched architecture is evaluated for multiple tasks. 

1) Parameter Settings for searching: To reduce the search time, 100 classes are randomly chosen from the original 1000 classes of ImageNet to build a training set and resize the images to the lower resolution of 32$\times$32. The training set is split into two equal subsets, one for tuning the network parameters and the other for tuning the architecture parameters. A network of 16 initial channels is trained for 20 epochs. A standard SGD optimizer with a momentum of 0.9 and a weight decay of 0.0003 is used to optimize the network parameters $w$. The initial learning rate is 0.025, which decays to zero following the cosine rule. Zero initialization is used for the architecture parameter $\alpha$, which implies an equal amount of attention over all possible operations. The Adam \cite{84} optimizer is used for the architecture parameters $\alpha$, with a learning rate of 0.0003 and a weight decay of 0.001.

2) Parameter Settings for fine-tuning: The CIFAR10 training images are randomly split into two groups, each containing 25 K images. One group is used for tuning the network parameters and the other for tuning the architecture parameters. The architecture is fine-tuned for 50 epochs.  A standard SGD optimizer is used to optimize the network parameters $w$ with an initial learning rate of 0.025, a momentum of 0.9, and a weight decay of 0.0003. The architectural parameters obtained in the search process are used to initialize the architecture parameter $\alpha$. The Adam \cite{84} optimizer is used for $\alpha$, with a learning rate of 0.0001 and a weight decay of 0.001. 

\begin{figure}[t]
	\centering
	\includegraphics[width=1\columnwidth]{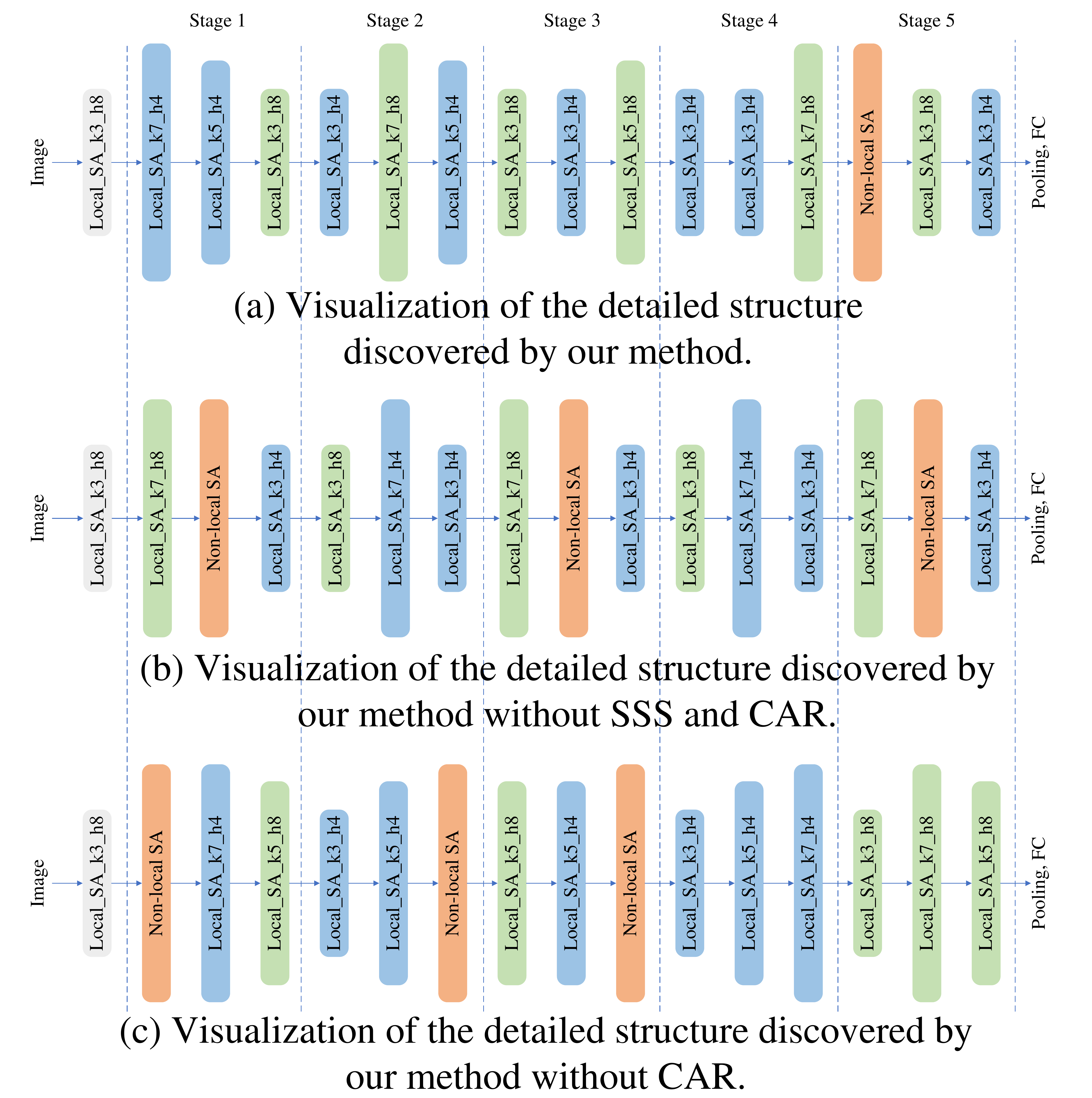} 
	\caption{ Visualizations of several searched architectures. The SSS refers to the stage-wise search space and CAR is the proposed search algorithm using the context auto-regression. Rounded rectangle boxes are used to denote blocks for each layer. The definition of the operations is described in \Cref{table2}. The first grey block ''Local SA\_k3\_h8'' is fixed. Different colors denote the types of the searched operation; orange for non-local self-attention with green and blue for local self-attention with different numbers of heads. Height is used to denote the spatial extent of the operation. }.
	\label{fig6}
\end{figure}

We ran four times with different random seeds to obtain four different models and selected the best model based on its validation performance. The proposed method takes approximately 2 GPU days to complete the search procedure on a single NVIDIA 1080Ti GPU. The discovered network is shown in \Cref{fig6}(a).


\subsection{Evaluation on image classification}
The search model is evaluated on the standard CIFAR10 and CIFAR100 image classification datasets.  The network of 96 initial channels is trained from scratch for 500 epochs.  The standard translation and flipping data augmentation scheme is applied to these datasets. The SGD optimizer is used with a momentum of 0.9 and a weight decay of 0.0004. The initial learning rate is 0.04, which decays to zero following the cosine rule.

The evaluation results and comparison with state-of-the-art approaches are summarized in \Cref{table3}. The proposed method achieved test errors of 3.44$\%$ and 20.68$\%$ on CIFAR-10 and CIFAR-100, respectively. As reported in Table 3, the experimental method outperforms recent strong baselines with significantly fewer computational resources. For example, it achieves superior performance compared to SOTA manually designed conv-based models. The test errors are significantly lower than the error rates achieved by the SENet architecture \cite{24}. With fewer parameters, it also achieves performance comparable to the automatically designed convolution-based models. In addition, it is compared with manual self-attention networks. The full self-attention models are trained using random initialization without extra pre-training and the proposed method outperformed them by a large margin.

\subsection{Transferability Validation on ImageNet}

To prove the transferability of our method, we transform the architecture learned on CIFAR-10 to a large and standard ImageNet dataset. The network of 112 initial channels is trained from scratch for 200 epochs using the cosine learning rate schedule with a base learning rate of 0.04. Standard data augmentation is applied to ImageNet, including random resized cropping, random horizontal flipping, and normalization. The network parameters are optimized using an SGD optimizer with a momentum of 0.9 and a weight decay of 3$\times$10-5. Label smoothing regularization is used [63] with a coefficient of 0.1 during training.

\Cref{table5} shows the quantitative results of ImageNet. On ImageNet, the network achieved top-1 and top-5 errors of 26.7$\%$ and 8.4$\%$, respectively. The experimental searched models are compared with state-of-the-art models, including manually designed convolution-based models, automatically designed convolution-based models, and manually designed self-attention models. The results indicate that the experimental model produces performance superior to these models. This demonstrates the transfer capability of the discovered self-attention architecture from a small dataset to a large dataset.

\begin{table}[]
	\centering
	\renewcommand{\arraystretch}{0.8}
	\renewcommand\tabcolsep{2pt}
	\begin{tabular}{cccc}
		\hline
		Method                    & Params & Top-1   Error & Top-5   Error \\ \hline
		Inception-V1   \cite{64}   & 6.6M   & 30.2\%        & 10.1\%        \\
		MobileNet-V1   \cite{65}   & 4.2M   & 29.4\%        & 10.5\%        \\
		MobileNet-V2    \cite{81}          & 3.4M   & 28.0\%        & 9.0\%         \\
		ShuffleNet   \cite{66}     & 5.0M   & 29.1\%        & 9.2\%         \\
		ResNet-18   \cite{39}      & 11.7M  & 28.5\%        & -             \\
		CondenseNet-V2 \cite{76} & 3.6M   & 28.1\%        & 9.7\%         \\ \hline
		NASNet-C        \cite{82}            & 4.9M   & 27.5\%        & 9.0\%         \\
		DARTS   \cite{7}          & 4.9M   & 26.9\%        & 9.0\%         \\
		SNAS      \cite{83}                & 4.3M   & 27.3\%        & 9.2\%         \\
		GDAS(FRC)     \cite{11}            & 4.4M   & 27.5\%        & 9.1\%         \\ \hline
		PVTv2\cite{79}             & 3.4M   & 29.5\%        & -             \\
		T2T-ViT-7 \cite{68}        & 4.3M   & 28.3\%        & -             \\
		DeiT-Ti    \cite{80}               & 5M     & 27.8\%        &               \\ \hline
		Our   method              & 4M     & 26.7\%        & 8.4\%         \\ \hline
	\end{tabular}
	\caption{Comparison with other architectures on ImageNet}
	\label{table5}
\end{table}

\subsection{Evaluation on Fine-grained Image Recognition}
To investigate further the transferability of the full-attention network searched on CIFAR-10, the capability of the discovered model is validated on the CUB-200-2011 dataset. In the experiments, the network structure searched on CIFAR-10 is applied as a feature extractor and combined with a classifier (e.g., fully-connected layers). For fair comparisons with other methods\cite{58,59}, each image is resized to 448 $\times$ 448. The network is trained for 200 epochs. The network parameters are optimized using an SGD optimizer with an initial learning rate of 0.02, a momentum of 0.9, and a weight decay of 0.0005.

The evaluation results summarized in \Cref{table6} reflect a detailed comparison with existing methods. ''Params'' denotes the parameters of the base model. Compared with other methods, our model achieves better performance by a large margin with fewer parameters. The results in Table \Cref{table6} demonstrate the transferability of architectures searched on the CIFAR-10 dataset and verify that our model is superior in more complex tasks.

\begin{table}[]
	\centering
	\renewcommand{\arraystretch}{0.8}
	\renewcommand\tabcolsep{2pt} 
	\begin{tabular}{cccc}
		\hline
		Method                     & Base Model             & Params & Accuracy \\ \hline
		Fine-tuned VGGNet\cite{53}   & VGG-19                 & 144M   & 77.80\%  \\
		CoSeg(+BBox)\cite{54}       & VGG-19                 & 144M   & 82.60\%  \\
		B-CNN\cite{55}              & VGG-16                 & 138M   & 84.00\%  \\
		CBP \cite{56}               & VGG-16                 & 138M   & 84.00\%  \\
		Fine-tuned ResNet\cite{39}  & ResNet-50              & 25.6M  & 84.10\%  \\
		LRBP \cite{57}             & VGG-16                 & 138M   & 84.20\%  \\
		FCAN   \cite{58}            & ResNet-50              & 25.6M  & 84.30\%  \\
		Kernel-Pooling\cite{59}     & ResNet-50              & 25.6M  & 84.70\%  \\ \hline
		Our method                 & \begin{tabular}[c]{@{}c@{}}Full-attention\\ Network\end{tabular}  & 3.82M  & 85.30\%  \\ \hline
	\end{tabular}
	\caption{Comparison of fine-grained image recognition methods on CUB-200-2011}
	\label{table6}
\end{table}

\subsection{Evaluation on zero-shot image retrieval}
In zero-shot image retrieval (ZSIR), the feature extractor is required to learn embedding from the seen classes and then to be capable of utilizing the learned knowledge to distinguish the unseen classes without any attributes or semantic information. Therefore, because the quality of features highly depends on it, the feature extractor is critically important. In this study, our discovered full-attention model is combined with a distance metric learning method, Proxy-Neighborhood Component Analysis \cite{68}, to validate the capability of the proposed model for zero-shot image retrieval. The Recall@K evaluation metric is computed for a direct comparison with previous methods on the CUB-200-2011 dataset. The first 100 classes are used as training data and the remaining 100 classes for evaluation. The inputs are resized to 256 $\times$ 256 pixels and then randomly cropped to 227 $\times$ 227 pixels. The network is trained for 50 epochs. The model is optimized using Adam \cite{84} with a learning rate of 4$\times$10-3. 

The evaluation results and a comparison with the state-of-the-art approaches are summarized in \Cref{table7}. From the table, our method improves the performance of zero-shot image retrieval by a large margin and achieves better performance than other methods. It is also notable that the full-attention networks obtained by our method have high transferability.

\begin{table}[]
	\centering
	\renewcommand{\arraystretch}{0.8}
	\renewcommand\tabcolsep{2pt} 
	\begin{tabular}{cccccc}
		\hline
		Method       & Backbone      & R@1 & R@2& R@4 & R@8 \\ \hline
		ProxyNCA\cite{68}   & InceptionBN     & 49.2 & 61.9& 67.9 & 72.4  \\
		MS\cite{69}       & InceptionBN       & 65.7 & 77& 86.3 & 91.2 \\
		HORDE\cite{70}       & InceptionBN       & 66.8 & 77.4& 85.1 & 91 \\
		XBM\cite{71}       & InceptionBN       & 65.8 & 75.9& 84 & 89.9 \\
		Margin\cite{72}       & ResNet-50       & 63.6 & 74.4& 83.1 & 90 \\
		NormSoftMax\cite{73}       & ResNet-50       & 65.3 & 76.7& 85.4 & 91.8 \\
		MIC \cite{74}      & ResNet-50      & 66.1 & 76.8& 85.6 & - \\  \hline
		Our method                 & \begin{tabular}[c]{@{}c@{}}Full-attention\\ Network\end{tabular}   & 68.4 & 78.5& 86 & 91.6  \\ \hline
	\end{tabular}
	\caption{Comparison of zero-shot image retrieval methods on CUB-200-2011.}
	\label{table7}
\end{table}

\begin{figure*}[t]
	\centering
	\includegraphics[width=0.9\linewidth]{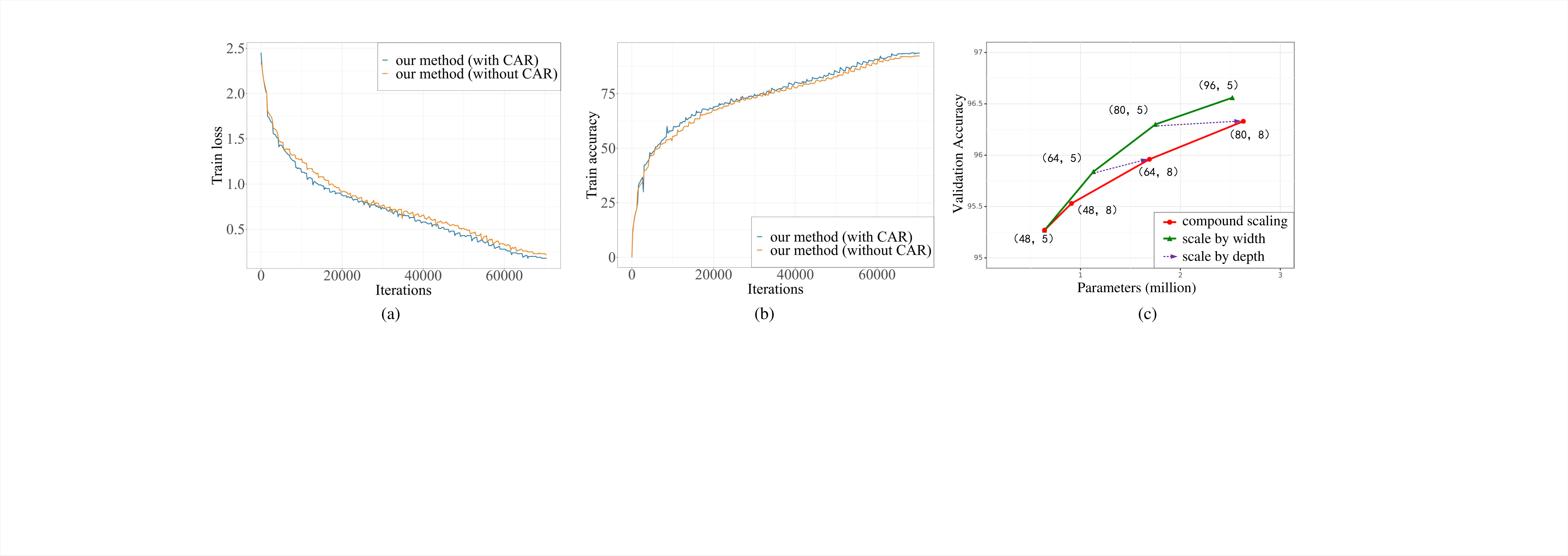} 
	\caption{ Results of ablation studies. (a) Loss curve with and without context auto-regression (CAR). (b) Accuracy curve with and without context auto-regression (CAR). (c) Scaling up a baseline model with different network width and depth. The first baseline network (C=48, S=5) has 48 initial channels and 5 stages (each stage has three intermediate layers).}.
	\label{fig7}
\end{figure*}

\subsection{Ablation Studies and Discussion}
In this section, we report on a series of ablation studies that validate the importance of the stage-wise search space as well as the proposed self-supervised search algorithm using context auto-regression incorporated in our method. All the architectures of 80 initial channels are trained for 500 epochs. \Cref{table8} shows the results of the ablation studies on CIFAR-10. SSS refers to the stage-wise search space, and CAR is the proposed search algorithm using context auto-regression. Then, we scale a baseline model with different network widths (w) and depths (d) and discuss the relationship between model dimensions and accuracy. 

1) Stage-wise Search Space (SSS) design: In order to study the impact of layer diversity brought by the proposed search space, a different search space is designed. In the searched model, the architectures of stages 1, 3, and 5 are identical as are the architectures of stages 2 and 4. The detailed structure discovered on CIFAR-10 using this search space design is shown in the \Cref{fig6}(b). As shown in \Cref{table8}, the network discovered by the method with our proposed stage-wise search space design exhibits high accuracy, highlighting the importance of the layer diversity throughout the network.

2) Search algorithm using Context Auto-Regression (CAR): \Cref{table8} demonstrates the efficacy of the proposed search algorithm using context auto-regression. ''Our method without CAR'' means searching on Cifar10 and the architecture parameters are initialized randomly. The results show that the searched model with context auto-regression exhibits high accuracy. The detailed structure discovered on CIFAR-10 by the method without context auto-regression is shown in \Cref{fig6}(c). We further show the loss and accuracy curves during the search process in \Cref{fig7}(a) and \Cref{fig7}(b). The figure demonstrates the efficacy of the proposed method. Owing to the initialized architecture parameters obtained through context auto-regression, our search network exhibits fast convergence and high training accuracy. 

3) Model scaling: \Cref{fig7}(c) shows our study on scaling a baseline model with different depths and widths. The scaling network width is commonly used for small-size models. The general trend is that our models perform better as the number of initial channels increases. Moreover, we attempt to repeat the structure of stage1, stage3, and stage5 once and obtain a deeper network with eight stages. When the network is narrow, the accuracy improves as the depth increases. However, the accuracy gains saturate when the model is wider. The intuition is that a deeper model can capture richer and more complex features. However, deeper networks are also more difficult to train due to the vanishing gradient problem. As shown in \Cref{fig7}(c), when the initial width is 96 and the stage is 5, the model achieves the highest accuracy.
\begin{table}[]
	\centering
	\renewcommand{\arraystretch}{0.8}
	\renewcommand\tabcolsep{3pt} 
	\begin{tabular}{ccccc}
		\hline
		Architecture                     & SSS & CAR & Parameters & Error  \\\hline
		\begin{tabular}[c]{@{}c@{}}Our method without\\  SSS and CAR\end{tabular} & $\times$   & $\times$   & 1.76M      & 4.38\% \\
		Our   method without CAR         & $\checkmark$   & $\times$   & 1.74M      & 4.08\% \\
		Our   method                     & $\checkmark$   & $\checkmark$   & 1.75M      & 3.7\% \\ \hline
	\end{tabular}
	\caption{Ablation studies on Cifar-10. SSS means the stage-wise search space. CAR means the search algorithm using the context auto-regression.}
	\label{table8}
\end{table}

\section{Conclusion}
In this article, we presented a full-attention based neural architecture search algorithm. A stage-wise search space was constructed to allow the search algorithm to flexibly explore various self-attention operations for different layers of the network. To extract global feature, a self-supervised search method using context auto-regression was proposed to learn full-attention architecture representations. To verify the effectiveness, we applied our method to image classification, fine-grained image recognition and zero-shot image retrieval. The extensive experiments demonstrated that the full-attention model discovered by our method drastically reduced the model parameters while achieving excellent model accuracies.

\label{sec:reference}
\bibliographystyle{IEEEtran}
\bibliography{reference}

\end{document}